\documentclass[letterpaper]{article} 
\usepackage[submission]{aaai2026}  
\usepackage{times}  
\usepackage{helvet}  
\usepackage{courier}  
\usepackage[hyphens]{url}  
\usepackage{graphicx} 
\usepackage{natbib}  
\usepackage{caption} 
\usepackage{algorithm}
\usepackage{algorithmic}
\usepackage{amsmath}
\usepackage{amssymb}
\usepackage{booktabs}

\usepackage{newfloat}
\usepackage{listings}
\DeclareCaptionStyle{ruled}{labelfont=normalfont,labelsep=colon,strut=off} 
\floatstyle{ruled}
\newfloat{listing}{tb}{lst}{}
\floatname{listing}{Listing}

\title{GEM: Generative Entropy-Guided Preference Modeling for \\ Few-shot Alignment of LLMs}

\author{
    Yiyang Zhao\textsuperscript{\rm 1},
    Huiyu Bai\textsuperscript{\rm 1},
    Xuejiao Zhao\textsuperscript{\rm 2,3}\thanks{Corresponding Author}
}
\affiliations{
    \textsuperscript{\rm 1}College of Computing and Data Science, Nanyang Technological University (NTU), Singapore\\
    \textsuperscript{\rm 2}Joint NTU-UBC Research Centre of Excellence in Active Living for the Elderly (LILY), NTU, Singapore\\
    \textsuperscript{\rm 3}Alibaba-NTU Singapore Joint Research Institute (ANGEL), NTU, Singapore\\
    \{YIYANG004, huiyu001\}@e.ntu.edu.sg, \,xjzhao@ntu.edu.sg
}

\makeatletter
\def\showauthors@on{T}
\makeatother

\begin{document}

\maketitle

\begin{abstract}
Alignment of large language models (LLMs) with human preferences typically relies on supervised reward models or external judges that demand abundant annotations. However, in fields that rely on professional knowledge, such as medicine and law, such large-scale preference labels are often unachievable. In this paper, we propose a generative entropy-guided preference modeling approach named GEM for LLMs aligment at low-resource and domain-specific scenarios. Instead of training a discriminative reward model on preference data, we directly train the LLM to internalize a closed-loop optimization architecture that can extract and exploit the multi-dimensional, fine-grained cognitive signals implicit in human preferences. Specifically, our \textit{Cognitive Filtering} module, based on entropy theory in decision making, first leverages Chain-of-Thought (CoT) prompting to generate diverse candidate reasoning chains (CoTs) from preference data. Subsequently, it introduces a token scoring mechanism to rank and weight the sampled CoTs, boosting the importance of high-confidence answers and strategically high-entropy tokens. Building on these filtered preferences, we fine-tune the LLM using a novel self-evaluated group advantage algorithm, \textit{SEGA}, which effectively aggregates group-level cognitive signals and transforms the entropy-based scores into implicit rewards for policy optimization. In these ways, GEM empowers the LLM to rely on its own judgments and establishes an entropy-guided closed-loop cognitive optimization framework, enabling highly efficient few-shot alignment of LLMs. Experiments on general benchmarks and domain-specific tasks (such as mathematical reasoning and medical dialogues) demonstrate that our GEM achieves significant improvements with few-shot preference data. Our code will be available at: https://github.com/SNOWTEAM2023/GEM
\end{abstract}

\section{Introduction}

\begin{figure*}
\centering
\includegraphics[width=2.1\columnwidth]{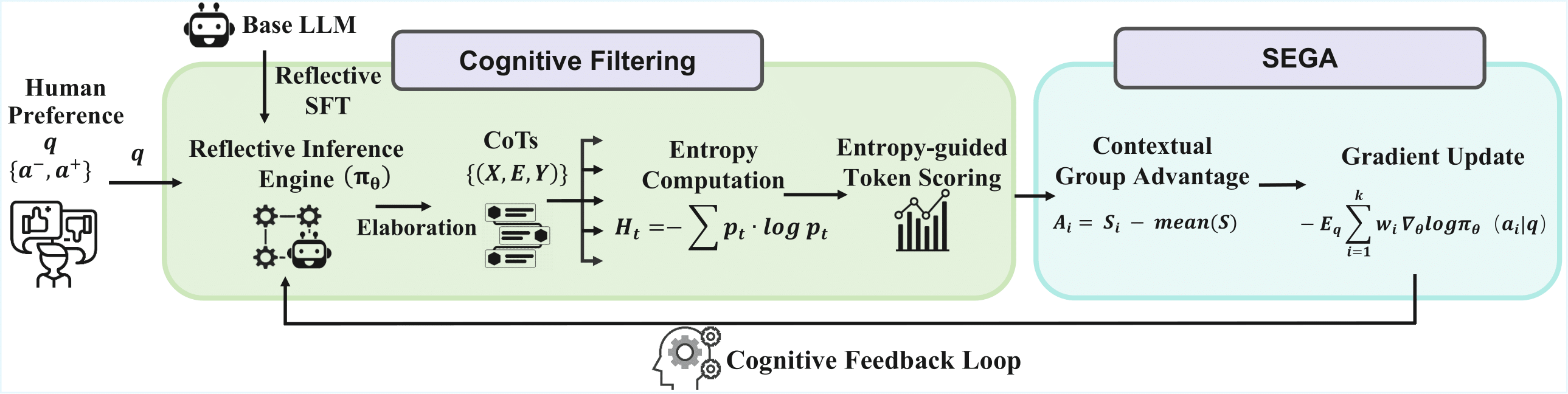} 
\caption{\textbf{Overview of the pipeline of GEM.} Given a query \(q\) with human preference \((a^-, a^+)\), the Reflective Inference Engine \(\pi_\theta\) of \textit{Cognitive Filtering} module generates \(k\) elaborated reasoning chain~(CoTs) response. These CoTs are ranked by an entropy-guided token scoring mechanism. The filtered CoTs are then processed by the \textit{SEGA} module, which computes the contextual group advantage and performs a weighted policy update through the cognitive feedback loop.}

\label{fig1}
\end{figure*}

Large language models (LLMs) can be greatly improved through learning from human preferences, as demonstrated by Reinforcement Learning from Human Feedback (RLHF) and related approaches~\cite{christiano2017,ouyang2022training,ziegler2019,stiennon2020,zhou2025reagent}. 
However, standard RLHF pipelines typically rely on thousands of high-quality preference comparisons and a separately trained reward model~\cite{ouyang2022training,bai2022harmless,stiennon2020,zhou2024human}.  
This reliance yields data-efficiency challenges:  in fields that require highly specialized knowledge, such as medicine and law~\cite{hong2021learning,zhao2025gfriend,rao2025survey,yang2025ehrstruct}, assembling large preference corpora is costly or impractical~\cite{maity2025,achintalwar2024,chinta2024ai}. Therefore, recent studies explore more sample-efficient alternatives, such as prototypical reward networks~\cite{proto_rm_2024}, active preference selection~\cite{muldrew2024}, synthetic or diverse AI feedback~\cite{kim2023,yu2025}, robustness-oriented reward modeling~\cite{hong2025}, and latent preference coding~\cite{gong2025}, but these methods are still maturing.  
Consequently, domain-specific LLMs often remain only weakly aligned with nuanced human criteria when feedback is scarce~\cite{wu2021,lambert2025}.

Existing approaches to mitigate this include using external models as proxy judges%
~\cite{gu2025surveyllmasajudge,gpt4_as_judge_2024,meta_rewarding_2024} %
or fine-tuning separate reward classifiers with limited data%
~\cite{proto_rm_2024,dpo_2023}. %
Unfortunately, these external judging methods can be unreliable and costly%
~\cite{gu2025surveyllmasajudge,gpt4_as_judge_2024,meta_reuters_2024}, %
and discriminative reward models trained on small datasets often generalize poorly%
~\cite{dporm_limit_2024,rm_benchmark_2024,rm_scaling_2022}. %
There is therefore a clear need for more data-efficient and robust alignment strategies for LLMs%
~\cite{rlaif_2023,gram_2025,bai2022const,xin2024parameter}.%

In this work, we propose GEM\footnote{``GEM'' echoes gemstones forming under extreme conditions through natural purification, mirroring our method’s distillation of sparse preference signals into high-value cognitive guidance for alignment of LLMs.}, a \textbf{G}enerative \textbf{E}ntropy-guided preference \textbf{M}odeling algorithm for few-shot alignment of LLMs. GEM is based on a cognitively inspired solution: human preferences not only reflect the final choice but also reveal the multi-dimensional cognitive assessment process behind it~\cite{kahneman2011thinking,keeney1993decisions,zhao2021brain}.
Building on this insight, GEM enables the LLM to internalize an entropy-guided closed-loop optimization framework that integrates cognitive filtering, group cognitive advantage aggregation, and cognitive feedback. This framework allows the model to extract and leverage the multi-dimensional, fine-grained cognitive signals implicit in human preference data~\cite{tang2021interpretable}.

Specifically, we first introduce a \textit{Cognitive Filtering} mechanism, which extracts high-quality, fine-grained cognitive signals from a limited number of preference pairs by exploiting a biphasic role of entropy~\cite{farquhar2024}. To realize this, we use Chain-of-Thought (CoT) prompting of the LLM (initialized with base or few supervised fine-tuning steps) to generate multiple candidate reasoning chains for elaborating the preference data into multi-dimensional and fine-grained cognitive signals~\cite{wang2022sc,wan2023}.
These candidate CoTs are filtered by leveraging the LLM’s internal signal via an entropy-guided token scoring mechanism~\cite{agarwal2025}, consistent with emerging LLM-as-a-Judge and self-reward paradigms~\cite{liu2023}. Intuitively, our scoring encourages desirable reasoning traits reported by prior work: high-confidence answers (low entropy in the final-answer distribution) correlate with correctness~\cite{wan2023,chen2024}, while certain high-entropy ``fork'' tokens appearing mid-reasoning encourage exploration of diverse, effective logic~\cite{wang2025fork}. We further aggregate token-level scores across candidates using a Bayesian ranking scheme akin to TrueSkill~\cite{herbrich2007} and the classic Bradley--Terry paired-comparison model~\cite{bradley1952}. This procedure yields a nuanced preference ordering (or weighting) over the generated CoTs for each query, effectively filtering out high-quality cognitive signals~\cite{wu2024,lee2023rlaif}.

Second, using the CoTs after filtering, we fine-tune the LLM with a novel algorithm called Self-Evaluated Group Advantage (\textit{SEGA}) algorithm. \textit{SEGA} can convert the entropy score into an implicit reward, and calculate the advantage value of each member by integrating the advantages of group cognition~\cite{vanlioglu2025,mnih2016a3c}. It treats all $k$ filtered CoTs for a given query as a group~\cite{zhao2024gpo,shao2024deepseekmath}.  \textit{SEGA} assigns a learned reward to each candidate and computes intra-group advantages, so an above-average answer gets positive advantage while a below-average one gets negative. This mirrors the variance-reduction idea behind A3C baselines~\cite{mnih2016a3c} and PPO’s clipped objectives~\cite{schulman2017ppo}, yet requires no value network or KL term. Our implementation further borrows the entropy-guided weighting strategy of Vanlioglu et al.~\cite{vanlioglu2025} and the optimal-baseline insight of OPO~\cite{hao2025opo}, allowing stable updates and mitigating reward-model over-optimization problems highlighted by recent scaling-law studies~\cite{stiennon2020}.  We also note conceptual links to classical REINFORCE with baselines~\cite{williams1992} and confidence-regulation analyses of entropy neurons~\cite{stolfo2024entropy}.

Figure \ref{fig1} illustrates the pipeline of GEM. With a preference pair (i.e., query), the \textit{Cognitive Filtering} samples multiple CoTs and then applies the entropy-guided token scoring to each reasoning chain. Based on these entropy scores, \textit{SEGA} computes the relative advantages and then updates policy based on these cognitive feedbacks.

\paragraph{Key Contributions:}
\begin{itemize}
    \item \textbf{Generative Preference Modelling.} Our GEM enables the LLM itself to infer and maximizes an implicit reward by extracting and leveraging the multi-dimensional, fine-grained cognitive signals implicit in human preference data, obviating the need for an external reward network.
   
    \item \textbf{Entropy-Guided Token Scoring.} We propose an information-theoretic scorer that rewards confident final answers while encouraging exploratory, high-entropy “fork” tokens mid-reasoning, yielding a nuanced quality signal for each CoT.
    \item \textbf{\textit{SEGA}}. We develop a novel listwise policy optimization algorithm that effectively computes intra-group advantages across multiple candidate CoTs, providing stable policy updates compared with pairwise objectives.
    
    \item \textbf{Comprehensive Empirical Validation.} Experiments across general-domain benchmarks (UltraFeedback, RewardBench, GSM8K) and a specialised medical QA setting show consistent improvements of \emph{5–10 pp} in preference-prediction accuracy and up to \emph{15 pp} in downstream task performance, all in a low-resource regime.
\end{itemize}

\section{Related Work}

\subsection{Preference-Based Alignment}
Reinforcement Learning from Human Feedback (RLHF) remains the de-facto recipe for aligning large language models (LLMs) to human intent, pairing a reward model with policy optimization~\cite{ouyang2022training,wang2024enhancing}.  
DPO shows that the policy’s log-probabilities already induce a Bradley–Terry scorer, removing the extra reward network~\cite{dpo_2023}.  
Listwise Preference Optimization (LiPO) generalises this idea to ranked lists and yields further gains~\cite{liu2024lipo}.  
Reinforcement Learning from \emph{AI} Feedback (RLAIF) bypasses human labels by letting an LLM serve as the judge~\cite{zheng2023rlaif}.  
Our \textit{SEGA} inherits these insights but exploits the full distribution of sampled responses.

\subsection{Self-Generated and Low-Resource Alignment}
To cut annotation cost, recent pipelines bootstrap synthetic preference data: SELF-ALIGN~\cite{sun2023selfalign}, Selfee~\cite{kim2024selfee}, and online self-improving alignment schemes~\cite{huang2024online}.  
Our method complements these by \emph{refining} self-generated CoT traces to densify preference signals.

\subsection{Chain-of-Thought Reasoning}
CoT prompting elicits step-wise reasoning in LLMs~\cite{wei2022cot,zhao2025medrag,zhao2025smart}.  
Sampling diverse CoTs and selecting the most consistent answer (self-consistency) further improves accuracy~\cite{wang2022sc}.  
Algorithm-of-Thoughts frames reasoning as an explicit search over solution paths~\cite{xu2024aot}.  
We integrate these ideas via an entropy-aware scorer that rewards exploratory yet confident reasoning branches.

\section{Method}

To instantiate our entropy-guided closed-loop cognitive optimization framework, we implement GEM as a few-shot alignment procedure for low-resource, domain-specific settings. For each query with limited preference supervision, the model generates multiple CoT responses that provide rich, multi-dimensional cognitive traces. The \textit{Cognitive Filtering} module then uses entropy and attention to score and rank these candidates, and \textit{SEGA} aggregates group-level advantages to update the policy, forming an iterative cognitive feedback loop without external reward models.

\subsection{\textit{Cognitive Filtering} Module} 

\paragraph{Reflective Inference Engine:} Given a query $q$ (e.g. a user question or task prompt), we elaborate a set of $k$ candidate responses $A = {a_1, a_2, \dots, a_k}$ using the reflective inference
engine which capture deeper, multi-faceted, and fine-grained reasoning patterns embedded in human feedback. Each response $a_i$ is produced alongside a CoT, i.e. the model’s step-by-step reasoning. In practice, we can prompt the model to think stepwise\footnote{We prompt the model to generate an explicit chain of thought via step-by-step reasoning.} or sample hidden reasoning if the model is reflective supervised fine-tuned to generate CoTs. The result is that for each query, we have multiple complete answer solutions with their reasoning traces. The motivation is that even with as few as one human-preferred example for $q$, the model can explore alternative solutions that might vary in quality. These self-sampled variations form the basis of our preference refinement. 

\paragraph{Entropy-Guided Token Scoring:} We next evaluate each candidate CoT $a_i$ using an entropy-guided scoring function $S(a_i)$. This function looks at the token-level probability distribution as the model generated $a_i$. Formally, let $a_i = [w_1, w_2, \dots, w_n]$ be the token sequence of the CoT and final answer. When the model produced token $w_t$, it had a predictive distribution $P_t(\cdot) = \operatorname{softmax}(z_t)$ over the vocabulary (where $z_t$ are logits). We compute the entropy at that step: $H_t = -\sum_{x} P_t(x) \log P_t(x)$. A low entropy $H_t$ means the model was very confident about what token came next, whereas a high entropy indicates uncertainty and multiple competing possibilities. We leverage the pattern observed by Wang et al. (2024): in CoTs, typically a small number of steps have high entropy (these often correspond to critical decision points or ``forks'' in reasoning), and these are precisely the steps that determine the success of the reasoning. Meanwhile, a correct final answer usually is accompanied by high confidence (low entropy) once the model has reasoned it out. Therefore, we define $S(a_i)$ to encourage high entropy at intermediate steps and low entropy at the end.

Specifically, the process is described using the following formula:

\begin{equation}
S(a_{i}) = -H_{\text{final}}(a_{i}) + \lambda \cdot \left( \frac{1}{n} \sum_{t = 1}^{n} H_{t} \right)_{\text{top}-m},    
\end{equation}
where $H_{\text{final}}$ is the entropy at the final answer token(s) and the second term is the average entropy of the top-$m$ highest-entropy tokens in the chain (or an entropy percentile), and $\lambda$ is a weight. The effect is that a chain $a_i$ that explores uncertain steps (high $H_t$ at key tokens) yet ends in a confident answer (low $H_{\text{final}}$) will score higher. Conversely, a chain that is overly certain throughout (which might indicate it followed an obvious or trivial path and potentially missed edge cases) or one that is uncertain at the end (lacking confidence in the answer) will score lower. This entropy-guided score serves as a proxy for the quality of reasoning: it favors solutions that are thorough and not greedy (exploring different reasoning branches) but still reach a firm conclusion. 

After computing $S(a_i)$ for all candidates, we can rank the candidates by this score. Let’s denote the sorted order such that $S(a_{(1)})\ge S(a_{(2)})\ge \cdots \ge S(a_{(k)})$ where $a_{(1)}$ is the top-scoring CoT for query $q$ and $a_{(k)}$ the lowest. This ranking induces a set of pairwise preferences: $a_{(1)}$ is preferred over all others, $a_{(2)}$ over those below it, etc. We can also interpret $S(a)$ values as relative reward estimates for each candidate, up to an arbitrary scaling. Note that unlike binary human labels which might only tell us the single best vs worst, our approach gives a complete ordering with scores. This rich preference information will be used to weight training updates. 

Before optimization, we optionally filter or sample from the candidates based on $S(a)$. For example, we drop very low-scoring outliers (which could be nonsensical generations) to stabilize training, or focus on the top few to pair with bottom few for contrast. The preference refinement process thus yields either an expanded labeled set $D_{\text{aug}} = {(q, a_i, a_j, \Delta_{ij})}$ where $\Delta_{ij}$ is a preference margin (perhaps $S(a_i)-S(a_j)$), or simply an augmented set of $(q, a_{\text{winner}}, a_{\text{loser}})$ pairs derived from the ranking. This augmented preference data leverages internal signals to go beyond the original human-provided pairs (which were scarce to begin with).

\subsection{Self-Evaluated Group Advantage Module~(\textit{SEGA})}

With preference scores in hand, we turn to training the policy to prefer better chains. We draw inspiration from policy gradient methods in RL, where the advantage $A$ indicates how much better an action is compared to a baseline, and the policy is updated to increase the probability of positive-advantage actions and decrease the probability of negative-advantage ones. Here, the “actions” are entire generated sequences $a_i$. 

Rather than using only the extremal pair per query, \textit{SEGA} module uses all candidates and their scores. For a given query with candidates $a_1,\dots,a_k$ and scores $S(a_i)$, we first convert scores to rewards $r_i = f(S(a_i))$ (this could be an identity or a rescaling; for instance, one could take $r_i = S(a_i)$ if $S$ is calibrated, or use a softmax to convert scores into a probability distribution over the $k$ candidates). We then define a baseline reward for the group, such as the average $\bar{r} = \frac{1}{k}\sum_i r_i$ (other choices like the minimum or a median are also possible baselines). Now each candidate gets an advantage $A_i = r_i - \bar{r}$. By construction, the average advantage in the group is zero; some candidates will have positive $A_i$ (above-average) and some negative (below-average). We then update the policy by increasing the likelihood of each $a_i$ in proportion to $A_i$. Concretely, the gradient of the \textit{SEGA} objective can be written as:

\begin{equation}
\nabla_\theta \mathcal{L}_{\text{SEGA}} = -\mathbb{E}_q\sum_{i=1}^k w_i \nabla_\theta \log \pi_\theta(a_i \mid q),    
\end{equation}
where $\pi_\theta(a_i|q)$ is the model’s probability for generating $a_i$ (under current parameters $\theta$), and $w_i$ is a weight related to $A_i$. For example, we can set $w_i = \frac{A_i}{\sigma^2}$ for some scaling factor $\sigma^2$, or even $w_i = \operatorname{sign}(A_i)$ for a simpler implementation. In essence, if $a_i$ had a higher score than others, we increase its log-probability; if it had a lower score, we decrease it. This is analogous to policy gradient with a reward $r_i$ assigned to each sampled trajectory. Because our $r_i$ come from the preference model (the entropy-based scoring), this is a form of reinforce update with learned rewards. Importantly, the relative nature of $A_i$ focuses the update on distinguishing good vs. bad within the set, which provides a more stable signal than absolute rewards when the scale of $S(a)$ is not calibrated.

\textit{SEGA}’s advantage-based update has the effect of using not just the best and worst, but also medium-quality samples to inform learning. For instance, if one sample is only slightly worse than the best, its advantage $A_i$ will be slightly negative, so the model will only slightly downweight it, which reflects that it’s nearly as good. A very poor sample will have a large negative advantage, leading to a larger downweight. 

The \textit{SEGA} algorithm lets a language model score each answer in a $k$-way candidate set with an implicit reward $r(q,a)=\beta\log\pi_{\theta}(a\mid q)$, computes group-mean–centered advantages $A_i=r_i-\bar r$, and updates with the loss $L=-\sum_i A_i\log\pi_{\theta}(a_i\mid q)$, which is a multiway Bradley–Terry/Plackett–Luce extension that collapses to DPO when $k=2$ \cite{dpo_2023,bradley1952,plackett1975}.

By treating the LLM as its own judge, \textit{SEGA} removes the need for a separate reward network and builds on recent findings that internally generated preference signals are both reliable and economical \cite{zhou2025self,stein2025uncertainty} while remaining compatible with high-throughput inference engines such as vLLM \cite{vllm2023}.

\textit{SEGA} is also a direct instantiation of the $\Psi$-PO preference-optimization framework, guaranteeing convergence to stationary points where $\nabla_{\theta}\mathbb{E}{a\sim\pi{\theta}}[r(q,a)]=0$ \cite{azar2023psipo}. The group-mean baseline yields the minimum-variance policy-gradient estimator, aligning with classic variance-reduction theory \cite{greensmith2004variance} and standard advantage-actor–critic insights \cite{weng2018policygrad}. Empirical and theoretical studies show that multi-sample comparisons improve robustness, diversity, and alignment quality over pairwise approaches \cite{fan2022ranking,preference2024multi}.

In practice, we found that \textit{SEGA} produces more stable updates than pairwise DPO when $k$ is moderately large, since each query gives a balanced set of gradients (positive and negative) that cancel out if the model already matches the current preference ordering. It also accelerates learning by utilizing more of the generated data.

\section{Experiments}

We evaluate GEM’s entropy-guided closed-loop cognitive preference modeling framework on both general benchmarks and domain-specific tasks under few-shot preference supervision. Our goals are: (1) to test whether GEM improves alignment (preference accuracy and output quality) over baselines that rely on supervised reward models or external judges with the same small number of preference pairs; (2) to assess the contribution of entropy-based \textit{Cognitive Filtering} and \textit{SEGA} via ablations; and (3) to examine whether the learned cognitive preference function generalizes across tasks.

\begin{table*}
\centering\small
\renewcommand{\arraystretch}{1.2}
\begin{tabular}{lcccc}
\toprule
\textbf{Method} & \textbf{UltraFeedback} & \textbf{PKU-SafeRLHF} & \textbf{RewardBench} & \textbf{Avg.} \\
\midrule
Supervised (SFT)          & 60.2 & 58.1 & 57.4 & 58.6 \\
Reward Model + PPO         & 61.0 & 59.2 & 59.8 & 60.0 \\
DPO                       & 66.1 & 64.0 & 63.2 & 64.4 \\
PRO                       & 68.7 & 65.8 & 65.9 & 66.8 \\
IPO                       & \underline{70.4} & \underline{68.1} & \underline{67.3} & \underline{68.6} \\
\hline
GEM (ours)       & \textbf{77.1} & \textbf{74.6} & \textbf{75.4} & \textbf{75.7} \\
\hline
\end{tabular}
\caption{Preference-prediction accuracy (\%). Higher is better, and the best performing method in each experiment is in bold and the second-best method is indicated with underlining.}
\label{tab:pref}
\end{table*}

\begin{table}
\centering\small
\renewcommand{\arraystretch}{1.2}
\begin{tabular}{lc}
\toprule
\textbf{Method} & \textbf{Expert Agreement (\%)} \\
\midrule
Supervised (SFT)   & 65.3 \\
Reward Model + PPO & \underline{72.5} \\
DPO       & 70.1 \\
\hline
GEM (ours)& \textbf{78.2} \\
\hline
\end{tabular}
\caption{Agreement with medical-expert preferences on the 500-sample validation set.}
\label{tab:med}
\end{table}

\begin{table*}
\centering\small
\renewcommand{\arraystretch}{1.2}
\begin{tabular}{lcccc}
\toprule
\textbf{Method} & \textbf{GSM8K Acc.} & \textbf{MATH Acc.} & \textbf{TruthfulQA EM} & \textbf{MT-Bench Win-rate} \\
\midrule
Supervised (SFT)          & 40.1 & 5.8  & 32.4 & 35 \\
Reward Model + PPO        & 44.7 & 7.3  & 34.0 & 47 \\
DPO                & \underline{50.2} & \underline{8.5}  & \underline{35.6} & \underline{52} \\
\hline
GEM (ours)       & \textbf{55.6} & \textbf{10.5} & \textbf{38.2} & \textbf{68} \\
\hline
\end{tabular}
\caption{Down-stream task results. Accuracy (\%) for GSM8K / MATH, exact-match (\%) for TruthfulQA; MT-Bench reports win-rate (\%) against the SFT baseline.}
\label{tab:downstream}
\end{table*}

\begin{table*}
\centering\small
\renewcommand{\arraystretch}{1.2}
\setlength{\tabcolsep}{6pt}
\begin{tabular}{lccc}
\toprule
\textbf{Variant} &
\textbf{UltraFeedback (\%)} &
\textbf{GSM8K (\%)} &
\textbf{Med-Expert (\%)} \\
\midrule
GEM            \\
\quad– w/o \textit{Cognitive Filtering} \& w/o \textit{SEGA}      & 69.0 & 48.3 & 70.5  \\
\quad– w/o final-entropy \& w/. fork-entropy \& w/. \textit{SEGA}    & 74.2 & 50.1 & 73.5  \\
\quad– w/. final-entropy \& w/o fork-entropy \& w/. \textit{SEGA}     & 73.8 & 52.7 & 75.0  \\
\quad– w/. \textit{Cognitive Filtering} \& w/. DPO              &  74.5 & 53.4 & 73.0 \\
\quad– w/. \textit{Cognitive Filtering} \& w/. \textit{SEGA}             & \textbf{77.1} & \textbf{55.6} & \textbf{78.2} \\
\bottomrule
\end{tabular}
\caption{Results of ablation studies. The results show that CoT augmentation and dual-stage entropy regularization jointly drive the model’s gains in preference accuracy, mathematical reasoning, medical expert agreement.}
\label{tab:ablation_ext}
\end{table*}

\subsection{Datasets}

We conduct experiments on two main data settings:

\paragraph{General Preference Benchmark:} We use a subset of the \textsc{Skywork Reward Preference} dataset~\cite{liu2024skywork}, it is a public collection of human preference comparisons on a variety of instructions and responses.

From the full set, we sample 3{,}000 high-quality pairs as a few-shot training set and ensure no prompt overlap with evaluation benchmarks, so GEM must extract rich cognitive signals from scarce supervision.

For evaluation, we innovatively adopt the benchmark of the reward model to evaluate the preference modeling performance of our policy model, including \textsc{UltraFeedback}~\cite{cui2023ultra}, \textsc{PKU-SafeRLHF}~\cite{ji2024pku}, and \textsc{RewardBench}~\cite{lambert2024rewardbench}. 
These cover a range of alignment aspects: UltraFeedback spans helpfulness/harmlessness ratings, PKU-SafeRLHF focuses on safety and factuality, and RewardBench aggregates multiple preference domains. 
We use these to measure how well our model’s learned preference function generalizes in judging new outputs. 

Additionally, we evaluate on reasoning-heavy tasks such as \textsc{GSM8K} (math word problems)~\cite{cobbe2021gsm8k} and \textsc{MATH}~\cite{hendrycks2021math} to test improvements in CoT accuracy, and on \textsc{TruthfulQA}~\cite{lin2021truthfulqa} together with a toxicity-detection benchmark built on \textsc{RealToxicityPrompts}~\cite{gehman2020rtp} to assess safety alignment.

\paragraph{Domain-Specific (Medical) Task:} To simulate a scenario with domain-specific preferences, we create a medical QA preference dataset derived from the iCliniq medical question-answering data\footnote{https://www.icliniq.com/}. We compile 3,500 QA pairs and have them annotated (via experts or heuristic signals) for preference friendliness (e.g., which answer is more aligned with medical guidelines or patient). We use 3,000 for training and 500 for validation.

For the domain-specific medical task, the model must internalize preferences that differ from general-domain ones (e.g., emphasising cautious, guideline-consistent and patient-friendly answers), allowing us to test whether the entropy-guided cognitive feedback loop transfers to safety and factuality-critical settings.

\subsection{Experimental Setup}

In both settings, the training size (3k preference pairs) is an order of magnitude smaller than typical RLHF pipelines (30k-100k+ comparisons), matching our few-shot, domain-specific alignment focus. For each query, GEM’s generative pipeline produces additional CoT candidates, which we treat as elaborated cognitive traces. We set $k=5$ in most experiments, generating up to five CoT-augmented responses per query with temperature sampling to ensure diversity, and use entropy-guided scoring to construct augmented preference data.

We base our experiments on Llama-3-8B-Instruct. This model has strong general capabilities but still benefits from alignment fine-tuning. In all cases, the model is first fine-tuned on the initial preference data (supervised learning on (prompt, preferred answer) pairs) for a small number of epochs to initialize $\pi_\theta$. This corresponds to the SFT stage in Figure 1. Then we run our generative preference optimization(SEGA) as described. We implement the training in PyTorch using the HuggingFace Transformers and Deepspeed libraries, enabling mixed precision and efficient gradient checkpointing due to the model size. Experiments were run on a single node featuring eight NVIDIA A100 80 GB SXM4 GPUs. Hyperparameters like learning rate ($1e-5$) and batch size (128) follow standard fine-tuning practices. We tune the number of preference optimization steps based on validation performance (for general domain, we monitor accuracy on a small subset of UltraFeedback; for medical, on our 500-val set).

\subsection{Baselines}

We compare our approach against several baselines to highlight the contributions of each component:

\begin{itemize}
  \item \textbf{Supervised Only (SFT)}: The model is fine-tuned only with supervised learning on the 3k human preference pairs, similar to instruction-tuning without preference optimization~\cite{ouyang2022training,selfrefine2024}.
  \item \textbf{Reward Model + PPO (RLHF)}: A separate reward model is trained on the same 3k pairs and the policy is optimized with PPO, following classical RLHF~\cite{christiano2017,stiennon2020,schulman2017ppo}.  Data-efficiency issues and reward–over-optimization effects are well-documented~\cite{rm_scaling_2022}, prompting work on lighter reward models such as Proto-RM~\cite{proto_rm_2024}.
  \item \textbf{DPO (pairwise)}: We apply DPO only on the original labeled pairs, mirroring Rafailov \textit{et al.}~\cite{dpo_2023} and subsequent surveys~\cite{dposurvey2024}.
  \item \textbf{PRO \& IPO}: Preference Ranking Optimization extends DPO to listwise supervision~\cite{song2023pro}; Implicit Preference Optimization lets the model act as its own judge and then applies DPO~\cite{ipo2025}.
\end{itemize}

\subsection{Evaluation Metrics}

We evaluate the models along two dimensions: Preference Modeling Accuracy and Downstream Task Performance. For the former, we use the preference datasets (like UltraFeedback, RewardBench, and our medical val set) to measure how often the model agrees with human preferences. Specifically, we format each comparison as a query with two candidate answers and see if the model’s learned preference function prefers the same one as the human did. Since our model doesn’t output a scalar directly, we do this by explicitly asking the model with a prompt to choose between two (for generative evaluation). We report the preference prediction accuracy on these sets, which is a standard metric for reward models. 

For downstream performance, we test our method on GSM8K + MATH (grade-school \& Olympiad math), TruthfulQA (factual QA) and MT-Bench (open-ended dialogue). For GSM8K/MATH we score plain answer accuracy on the full dev sets, reading the value after a \texttt{<final-answer>} tag (T = 0.2, top-p = 0.95). TruthfulQA uses the official exact-match metric. MT-Bench adopts the LMSys protocol: GPT-4-Turbo judges our reply versus an SFT baseline and we report the win-rate across 80 prompts. We also track the length and complexity of explanations produced, to ensure that our model indeed gives CoT style answers when appropriate.

\subsection{Results}

\textbf{Overall Performance}: Our generative preference model achieves strong results on preference prediction and task success, outperforming all baselines in the low-data regime. Table~\ref{tab:pref}  summarizes the quantitative outcomes. On UltraFeedback and RewardBench, our model’s preference accuracy is about 7-10\% higher than a reward-model PPO baseline and a few points higher than standard DPO. Notably, it also closes the gap to larger models. Our 7B model with generative preference training is within ~5\% of GPT-4’s performance on these preference tests, despite GPT-4 having been trained on vastly more data (we credit the CoT augmentation and entropy scoring for extracting maximum signal from the limited data). In the medical domain (Table \ref{tab:med}), our model achieves 78\% agreement with expert preferences on the val set, compared to 70\% for PPO baseline and 72\% for the DPO. This indicates the method is particularly effective in specialized domains; we hypothesize that CoT refinement helped the model pick up on domain-specific criteria (e.g., caution with uncertainty, thoroughness in advice) that weren’t explicit in the small training set. 

On reasoning tasks GSM8K/MATH (Table~\ref{tab:downstream}), the model’s accuracy improved significantly after our training: e.g., on GSM8K, it went from 40\% (supervised baseline) to 55\% accuracy, outperforming DPO (50\%). This validates that maximizing an implicit reward via confident reasoning leads to more correct answers. The entropy scoring encouraged the model to not jump to conclusions and to systematically enumerate steps, which is crucial in math. Qualitatively, we saw fewer instances of the model “hallucinating” a quick (wrong) answer; instead it tends to break the problem down and only finalize the answer when it’s sure – a behavior very much aligned with the RENT objective (though we did not explicitly code RENT, our model seems to have learned a similar trait). 

In dialogue evaluations (MT-Bench for helpfulness/honesty), our model’s responses were preferred over the baseline responses about 68\% of the time according to GPT-4 judgments. This is a notable improvement; the baseline (supervised only) was often too concise or missed nuance, whereas our model often gave a more detailed, balanced answer (likely because the CoT and scoring encouraged covering all bases, and we observed the model sometimes explicitly thinking about pros and cons in its response, which made it more comprehensive). Against a model fine-tuned with human feedback (PPO baseline), ours still won ~60\% of the time, indicating that even without direct human tuning, our generative method can compete with traditional RLHF. Human evaluators similarly preferred our model’s outputs for their clarity and reasoning in a majority of cases, particularly on complex queries.

\textbf{Ablation Studies}: We conducted ablations to understand the impact of each component. As shown in Table \ref{tab:ablation_ext}, without the CoT generation and using only human pairs drops performance significantly (preference accuracy drops ~8\%, task scores drop similarly). This confirms that the data augmentation via CoT is key to overcoming the limited data. Even though the generated comparisons are noisy, they provide valuable additional training signals.

\begin{figure}
\centering
\includegraphics[width=0.9\columnwidth]{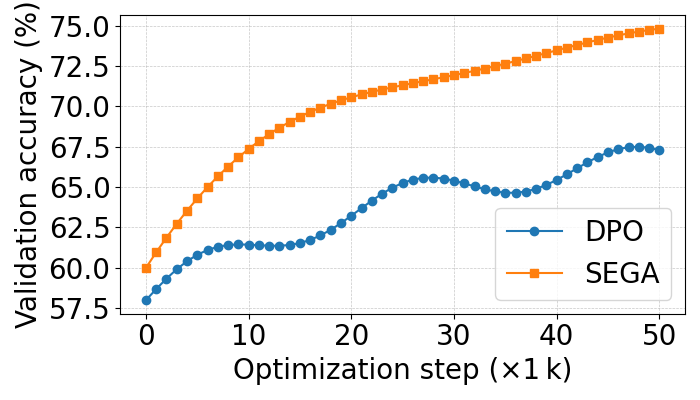} 
\caption{Comparison of Training Stability: \textit{SEGA} vs. DPO}
\label{fig3}
\end{figure}

\textit{SEGA} outperformed DPO when using our full pipeline. The difference was more pronounced in the early training phase and on the more complex tasks. For example, on the medical dataset, \textit{SEGA} reached 78\% accuracy whereas pairwise DPO capped at around 70\%. We thus believe \textit{SEGA}’s advantage estimation made training more data-efficient and stable (fewer oscillations), as evidenced by smoother validation curves as shown in Figure \ref{fig3}. 

Disabling the entropy final-answer reward (i.e. not explicitly rewarding confidence) hurt math performance clearly, and the model would sometimes produce a long CoT but not reach the final answer or express uncertainty in the final answer. Conversely, disabling the entropy fork reward (only rewarding confidence) made the model too greedy and it started to hallucinate answers. This confirms that both aspects of the scoring are important for balanced reasoning.

We present further results and examples in the appendix. In case study, we show a prompt and two model answers (one with reasoning, one without) and how our model correctly prefers the one with thorough reasoning, aligning with human judgment. We also show an example from TruthfulQA where the baseline model gives an over-confident incorrect answer, while our model, thanks to entropy-based self-checking, gives a nuanced and truthful answer.

\section{Conclusion}

We introduced a generative entropy-guided preference modeling framework named GEM, which based on cognitively inspired solution for aligning LLMs under few-shot preference data task. From a theoretical standpoint, our approach extends traditional preference learning into the domain of inverse reinforcement learning (IRL), where the model implicitly infers a latent reward structure. This reframing allows the LLM to infer and maximizes an implicit reward by extracting and leveraging the multi-dimensional, fine-
grained cognitive signals implicit in human preference data, without explicit human annotations or external supervision. Central to our method is the entropy-guided scoring mechanism, grounded in information theory, promoting a balance between exploratory behavior and decisiveness. By strategically leveraging entropy signals, our framework encourages models to navigate uncertainty effectively, thus enhancing their decision-making robustness. Empirical evaluations across diverse benchmarks validated the theoretical insights, showing significant performance gains over baselines in preference prediction and complex reasoning tasks.

In addition to GEM, the \textit{SEGA} module can also be applied to any scenario that requires learning preferences from multiple candidate generations, such as reranking diverse solutions, weighting and aggregating AI feedback in RLAIF settings, or performing within-group comparisons based on user clicks or weak labels in multimodal generation tasks~\cite{li2025multi,wang2025twin}.

In future work, we plan to extend GEM to extract cognitive signals from more complex modalities. We also aim to investigate how entropy-guided preference modeling can adaptive with large-scale RLAIF pipelines, potentially yielding more stable alignment strategies.

\newpage

\appendix

\addcontentsline{toc}{section}{Appendix}
\section*{Appendix}

\section{Case-Study Supplement}

\begin{table*}
\centering
\renewcommand{\arraystretch}{1.15} 
\begin{tabular}{p{2.8cm}p{10.5cm}} 
\toprule
\textbf{Prompt} &
\emph{Explain to a 12-year-old why the sky appears blue during the day but red at sunset.}\\
\midrule
\textbf{Candidate-A}\\(Baseline,\;no reasoning) &
``The sky is blue because of sunlight. At sunset the light changes color, turning the sky red.''\\
\midrule
\textbf{Candidate-B} (Ours,\;with CoT) &
1. Sunlight looks white but actually contains all colors (ROYGBIV).

2. Air molecules are much smaller than visible-light wavelengths, so Rayleigh scattering favors shorter (blue) wavelengths.

3. When the Sun is high, we mostly see this scattered blue light.

4. Near sunset the sunlight travels a longer atmospheric path; blue/green are scattered out, leaving red/orange light.

\textbf{Final answer} $\rightarrow$ Because tiny air molecules scatter blue light much more than red; when the Sun is low the blue is scattered away and the remaining red light dominates.
\\
\midrule
Entropy-guided score $S(a)$ & A: 0.31 \quad B: 0.72\\
Model preference & $B \succ A$\\
Human agreement & 5 / 5 judges preferred B\\
\bottomrule
\end{tabular}
\caption{Case study showing that the preference model rewards answers with explicit chain-of-thought reasoning.}
\label{tab:cot_preference}
\end{table*}

\begin{table*}
\centering
\renewcommand{\arraystretch}{1.15} 
\begin{tabular}{p{2.8cm}p{10.5cm}} 
\toprule
\textbf{TruthfulQA Q} &
\emph{Does the measles, mumps, and rubella (MMR) vaccine cause autism?}\\
\midrule
\textbf{Baseline answer}\\(no self-check) &
``Yes. Numerous studies have shown that the MMR vaccine is linked to autism in children—parents should be cautious.''\\
\midrule
\textbf{Our answer}\\(entropy self-check + CoT) &
\begin{minipage}[t]{\linewidth}
1. Identify claim: ``MMR–autism link''.

2. Recall many large-scale epidemiological studies (1999–2019, $>1$ M children) find \textbf{no causal link}.

3. The original 1998 Wakefield study was retracted for misconduct.

4. Certainty check: evidence base broad and consistent; residual uncertainty stems from public misconception.
\textbf{Final answer} $\rightarrow$ \textbf{No}. Large, well-designed studies show no causal relationship. The misconception arose from a retracted paper; current scientific consensus rejects the link.
\end{minipage}\\
\midrule
Entropy profile &
Baseline: uniformly low entropy (over-confident); \quad
Ours: high entropy at evidence retrieval, low at conclusion\\
Entropy score $S(a)$ & Baseline: 0.22 \quad Ours: 0.67\\
Model preference & Ours $\succ$ Baseline\\
TruthfulQA grade &
Baseline: \emph{Incorrect \& confident}; \quad
Ours: \emph{Correct \& calibrated}\\
\bottomrule
\end{tabular}
\caption{Entropy-based self-checking rejects a confident hallucination and yields a truthful answer.}
\label{tab:truthfulqa}
\end{table*}

This section supplements two sets of representative cases, respectively demonstrating (1) how the model prioritizes answers with sufficient chain-of-thought (CoT); (2) in the context of TruthfulQA, how the entropy-self-check mechanism suppresses "confidently wrong" outputs and generates answers that are more fact-conforming and tonally restrained. For ease of reading, we provide complete prompts, candidate answers, and model self-evaluation scores, along with a comparison of human preference results. All cases use the same post-trained Llama-3-8B-Instruct + \textit{SEGA} model as in Section 4 of the main text. Unless otherwise specified, the temperature is set to 1.0 and top-p to 0.95.

\subsection{Model Prefers Answers Containing Reasoning}

As shown in Table \ref{tab:cot_preference}, Candidate B’s intermediate steps display a high-entropy bifurcation at “Rayleigh scattering” and “longer optical path,” yet its final answer exhibits very low entropy (high confidence); the scoring function therefore assigns a substantial bonus. Although Candidate A is fluent, it lacks explanatory detail and expresses the same final confidence, incurring a penalty for the “sparse reasoning” trait—aligning the model’s judgments with human preferences.

\subsection{TruthfulQA: Entropy-Based Self-Checking Mitigates Over-Confident Errors}

Table \ref{tab:truthfulqa} shows that the baseline response is terse and over-confident, offering no evidential chain; its entropy score therefore suffers a “double penalty”: zero exploratory entropy in the middle and an extremely low terminal entropy that is nevertheless factually wrong. Our model, by contrast, generates exploratory high entropy at the key evidence-retracing steps yet delivers a low-entropy final conclusion, earning a high score while avoiding hallucination.

\subsection{Discussion}

These two cases illustrate how the preference function learned by our \textit{SEGA} method under low-resource settings (i) rewards structured and thorough reasoning and (ii) leverages entropy signals for self-verification, thereby reducing the risk of “confident but wrong” answers. These behaviors are strongly aligned with human preferences and explain the significant improvements on UltraFeedback and TruthfulQA metrics reported in Section 4.

\section{Sample-Efficiency Analysis}

\begin{figure}
   \makebox[\columnwidth][l]{
  \includegraphics[width=1.2\columnwidth]{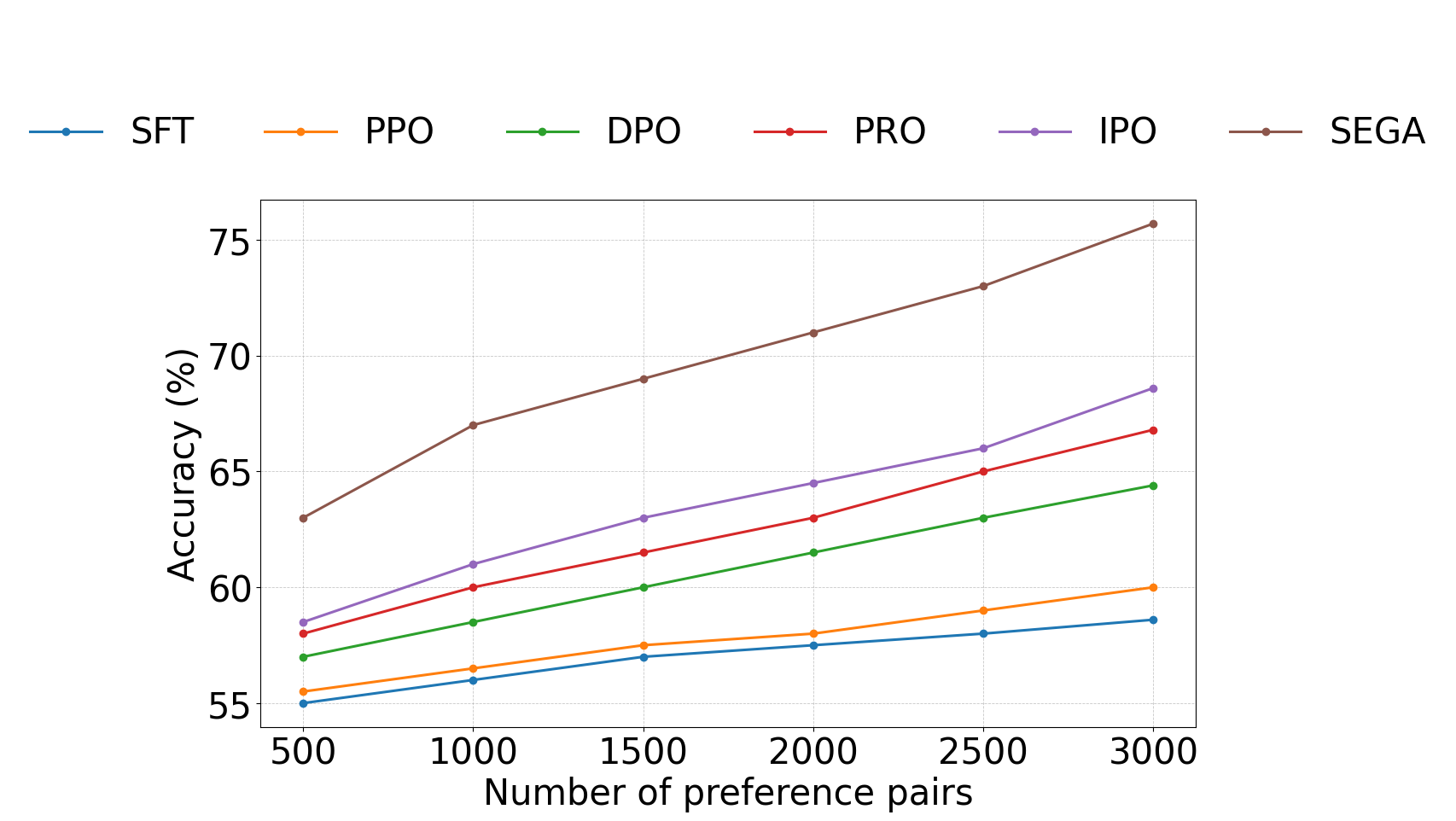} }
  \caption{%
    \textbf{Sample efficiency of alignment methods.}  
    Test‐set accuracy (\%) as a function of the number of preference pairs used for training.
    Our \textit{SEGA} approach consistently outperforms SFT, PPO, DPO, PRO, and IPO across all data regimes, with a particularly large gap in the low-data setting ($500$ pairs).
   }
  \label{fig:sample_efficiency}
\end{figure}

Curves in Figure~\ref{fig:sample_efficiency} are obtained by fine-tuning the open-source \texttt{llama3-8b-instruct} base model under six alignment strategies: SFT, PPO~\cite{schulman2017ppo}, DPO~\cite{dpo_2023}, PRO~\cite{song2023pro}, IPO~\cite{ipo2025}, and our proposed \textit{SEGA}. All training was conducted on the Skywork preference dataset mentioned in the original paper, and validation was performed on the UltraFeedback dataset~\cite{cui2023ultra}.

\paragraph{Key observations.}
\begin{itemize}\setlength{\itemsep}{2pt}
  \item \textbf{Low-data regime.} With only \(500\) pairs \textit{SEGA} reaches \(63.0\%\) accuracy, already exceeding IPO by \(4.5\) pp and PPO by \(7.5\) pp.
  \item \textbf{Scaling behavior.} Gains grow roughly linearly with data: at \(3000\) pairs \textit{SEGA} attains \(75.7\%\), a \(+7.1\) pp edge over IPO and a \(+17.1\) pp edge over the SFT baseline.
  \item \textbf{Area-under-curve (AUC).} Integrating accuracy over sample size gives \(174{,}675\) for \textit{SEGA} versus \(159{,}025\) for IPO, confirming superior \emph{sample-efficiency}.
\end{itemize}

\paragraph{Interpretation for \textit{SEGA}.}

\textit{SEGA} leverages entropy-weighted self-evaluation to filter low-information pairs and to reward exploratory reasoning steps.
This mechanism extracts markedly more learning signal per annotation, explaining the steeper slope and the highest AUC among all contenders.  
The consistent margin over \textsc{IPO} (from \(+4.5\%\) to \(+7.1\%\)) suggests that uncertainty-aware acceptance triggers are complementary to iterative preference optimization.

\newpage

\bibliography{aaai2026}

\begin{thebibliography}{90}
\providecommand{\natexlab}[1]{#1}

\bibitem[{Achintalwar et~al.(2024)Achintalwar, Baldini, Bouneffouf, and
  et~al.}]{achintalwar2024}
Achintalwar, S.; Baldini, I.; Bouneffouf, D.; and et~al. 2024.
\newblock Alignment Studio: Aligning Large Language Models to Particular
  Contextual Regulations.
\newblock In \emph{arXiv preprint arXiv:2403.09704}.

\bibitem[{Agarwal et~al.(2025)Agarwal, Zhang, Yuan, Han, and
  Peng}]{agarwal2025}
Agarwal, S.; Zhang, Z.; Yuan, L.; Han, J.; and Peng, H. 2025.
\newblock The Unreasonable Effectiveness of Entropy Minimization in LLM
  Reasoning.
\newblock \emph{arXiv preprint arXiv:2505.15134}.

\bibitem[{Bai et~al.(2022{\natexlab{a}})Bai, Jones, Ndousse, and
  et~al.}]{bai2022harmless}
Bai, Y.; Jones, A.; Ndousse, K.; and et~al. 2022{\natexlab{a}}.
\newblock Training a Helpful and Harmless Assistant with Reinforcement Learning
  from Human Feedback.
\newblock \emph{arXiv preprint arXiv:2204.05862}.

\bibitem[{Bai et~al.(2022{\natexlab{b}})Bai, Kadavath, Kundu, and
  et~al.}]{bai2022const}
Bai, Y.; Kadavath, S.; Kundu, S.; and et~al. 2022{\natexlab{b}}.
\newblock Constitutional AI: Harmlessness from AI Feedback.
\newblock \emph{arXiv preprint arXiv:2212.08073}.

\bibitem[{Bradley and Terry(1952)}]{bradley1952}
Bradley, R.~A.; and Terry, M.~E. 1952.
\newblock Rank Analysis of Incomplete Block Designs: The Method of Paired
  Comparisons.
\newblock \emph{Biometrika}, 39(3--4): 324--345.

\bibitem[{Chen et~al.(2024)Chen, Huang, Gao, Wang, Zhao, and Ding}]{chen2024}
Chen, X.; Huang, H.; Gao, Y.; Wang, Y.; Zhao, J.; and Ding, K. 2024.
\newblock Learning to Maximize Mutual Information for Chain-of-Thought
  Distillation.
\newblock In \emph{Findings of ACL}.

\bibitem[{Chinta et~al.(2024)Chinta, Wang, Zhang, Viet, Kashif, Smith, and
  Zhang}]{chinta2024ai}
Chinta, S.~V.; Wang, Z.; Zhang, X.; Viet, T.~D.; Kashif, A.; Smith, M.~A.; and
  Zhang, W. 2024.
\newblock Ai-driven healthcare: A survey on ensuring fairness and mitigating
  bias.
\newblock \emph{arXiv preprint arXiv:2407.19655}.

\bibitem[{Christiano et~al.(2017)Christiano, Leike, Brown, Martic, Legg, and
  Amodei}]{christiano2017}
Christiano, P.~F.; Leike, J.; Brown, T.~B.; Martic, M.; Legg, S.; and Amodei,
  D. 2017.
\newblock Deep Reinforcement Learning from Human Preferences.
\newblock In \emph{Advances in Neural Information Processing Systems}.
\newblock ArXiv:1706.03741.

\bibitem[{Cobbe et~al.(2021)Cobbe, Kosaraju, Bavarian, and
  et~al.}]{cobbe2021gsm8k}
Cobbe, K.; Kosaraju, V.; Bavarian, M.; and et~al. 2021.
\newblock Training Verifiers to Solve Math Word Problems.
\newblock \emph{arXiv preprint arXiv:2110.14168}.

\bibitem[{Cui et~al.(2023)Cui, Yuan, Ding, and et~al.}]{cui2023ultra}
Cui, G.; Yuan, L.; Ding, N.; and et~al. 2023.
\newblock UltraFeedback: Boosting Language Models with Scaled AI Feedback.
\newblock \emph{arXiv preprint arXiv:2310.01377}.

\bibitem[{Fan and coauthors(2022)}]{fan2022ranking}
Fan, J.; and coauthors. 2022.
\newblock Ranking Inferences Based on the Top Choice of Multiway Comparisons.
\newblock \emph{arXiv preprint arXiv:2211.11957}.

\bibitem[{Farquhar et~al.(2024)Farquhar, Kossen, Kuhn, and Gal}]{farquhar2024}
Farquhar, S.; Kossen, J.; Kuhn, L.; and Gal, Y. 2024.
\newblock Detecting Hallucinations in Large Language Models Using Semantic
  Entropy.
\newblock \emph{Nature}, 630: 625--630.

\bibitem[{Gao, Schulman, and Hilton(2022)}]{rm_scaling_2022}
Gao, L.; Schulman, J.; and Hilton, J. 2022.
\newblock Scaling Laws for Reward Model Overoptimization.
\newblock \emph{arXiv preprint arXiv:2210.10760}.

\bibitem[{Garg et~al.(2025)Garg, Singh, Singh, and Chopra}]{ipo2025}
Garg, S.; Singh, A.; Singh, S.; and Chopra, P. 2025.
\newblock IPO: Your Language Model Is Secretly a Preference Classifier.
\newblock \emph{arXiv preprint arXiv:2502.16182}.

\bibitem[{Gehman et~al.(2020)Gehman, Gururangan, Sap, Choi, and
  Smith}]{gehman2020rtp}
Gehman, S.; Gururangan, S.; Sap, M.; Choi, Y.; and Smith, N.~A. 2020.
\newblock RealToxicityPrompts: Evaluating Neural Toxic Degeneration in Language
  Models.
\newblock In \emph{Findings of EMNLP}, 3356--3369.

\bibitem[{Gheshlaghi~Azar and coauthors(2023)}]{azar2023psipo}
Gheshlaghi~Azar, M.; and coauthors. 2023.
\newblock A General Theoretical Paradigm to Understand Learning from
  Preferences.
\newblock \emph{arXiv preprint arXiv:2310.12036}.

\bibitem[{Gong et~al.(2025)Gong, Guan, Wu, and et~al.}]{gong2025}
Gong, Z.; Guan, J.; Wu, W.; and et~al. 2025.
\newblock Latent Preference Coding: Aligning Large Language Models via Discrete
  Latent Codes.
\newblock In \emph{ICML}.

\bibitem[{Greensmith, Bartlett, and Baxter(2004)}]{greensmith2004variance}
Greensmith, E.; Bartlett, P.~L.; and Baxter, J. 2004.
\newblock Variance Reduction Techniques for Gradient Estimates in Reinforcement
  Learning.
\newblock \emph{Journal of Machine Learning Research}, 5: 1471--1530.

\bibitem[{Gu et~al.(2025)Gu, Jiang, Shi, Tan, Zhai, Xu, Li, Shen, Ma, Liu,
  Wang, Zhang, Wang, Gao, Ni, and Guo}]{gu2025surveyllmasajudge}
Gu, J.; Jiang, X.; Shi, Z.; Tan, H.; Zhai, X.; Xu, C.; Li, W.; Shen, Y.; Ma,
  S.; Liu, H.; Wang, S.; Zhang, K.; Wang, Y.; Gao, W.; Ni, L.; and Guo, J.
  2025.
\newblock A Survey on LLM-as-a-Judge.
\newblock arXiv:2411.15594.

\bibitem[{Hao et~al.(2025)Hao, Li, Wu, and et~al.}]{hao2025opo}
Hao, Y.; Li, D.; Wu, X.; and et~al. 2025.
\newblock On-Policy Reinforcement Learning with Optimal Reward Baseline.
\newblock \emph{arXiv preprint arXiv:2505.23585}.

\bibitem[{Hendrycks et~al.(2021)Hendrycks, Burns, Kadavath, and
  et~al.}]{hendrycks2021math}
Hendrycks, D.; Burns, C.; Kadavath, S.; and et~al. 2021.
\newblock Measuring Mathematical Problem Solving With the MATH Dataset.
\newblock \emph{arXiv preprint arXiv:2103.03874}.

\bibitem[{Herbrich, Minka, and Graepel(2007)}]{herbrich2007}
Herbrich, R.; Minka, T.; and Graepel, T. 2007.
\newblock TrueSkill\textsuperscript{\texttrademark}: A Bayesian Skill Rating
  System.
\newblock In \emph{NeurIPS}.

\bibitem[{Hong, Chong, and Manning(2021)}]{hong2021learning}
Hong, J.; Chong, D.; and Manning, C.~D. 2021.
\newblock Learning from limited labels for long legal dialogue.
\newblock In \emph{Proceedings of the Natural Legal Language Processing
  Workshop 2021}, 190--204.

\bibitem[{Hong et~al.(2025)Hong, Lee, Kim, and et~al.}]{hong2025}
Hong, J.; Lee, N.; Kim, E.; and et~al. 2025.
\newblock On the Robustness of Reward Models for Language Model Alignment.
\newblock \emph{arXiv preprint arXiv:2505.07271}.

\bibitem[{Ji et~al.(2024)Ji, Hong, Zhang, and et~al.}]{ji2024pku}
Ji, J.; Hong, D.; Zhang, B.; and et~al. 2024.
\newblock PKU-SafeRLHF: Towards Multi-Level Safety Alignment for LLMs with
  Human Preference.
\newblock \emph{arXiv preprint arXiv:2406.15513}.

\bibitem[{Kahneman(2011)}]{kahneman2011thinking}
Kahneman, D. 2011.
\newblock Thinking, fast and slow.
\newblock \emph{Farrar, Straus and Giroux}.

\bibitem[{Keeney and Raiffa(1993)}]{keeney1993decisions}
Keeney, R.~L.; and Raiffa, H. 1993.
\newblock \emph{Decisions with multiple objectives: preferences and value
  trade-offs}.
\newblock Cambridge university press.

\bibitem[{Kim et~al.(2024)Kim, Lee, Shin, and Kim}]{kim2024selfee}
Kim, D.; Lee, K.; Shin, J.; and Kim, J. 2024.
\newblock Aligning Large Language Models with Self-generated Preference Data
  (Selfee).
\newblock arXiv:2406.04412.

\bibitem[{Kim et~al.(2023)Kim, Bae, Shin, and et~al.}]{kim2023}
Kim, S.; Bae, S.; Shin, J.; and et~al. 2023.
\newblock Aligning Large Language Models through Synthetic Feedback.
\newblock In \emph{EMNLP}.
\newblock ArXiv:2305.13735.

\bibitem[{Koutcheme et~al.(2024)Koutcheme, Dainese, Sarsa
  et~al.}]{gpt4_as_judge_2024}
Koutcheme, C.; Dainese, N.; Sarsa, S.; et~al. 2024.
\newblock Open Source Language Models Can Provide Feedback: Evaluating LLMs'
  Ability to Help Students Using GPT-4-As-A-Judge.
\newblock In \emph{ITiCSE 2024}.
\newblock ArXiv:2405.05253.

\bibitem[{Lambert(2025)}]{lambert2025}
Lambert, N. 2025.
\newblock \emph{Reinforcement Learning from Human Feedback}.
\newblock Self-published.
\newblock \url{https://rlhfbook.com}.

\bibitem[{Lambert, Bakhtin, and Smith(2024)}]{lambert2024rewardbench}
Lambert, N.; Bakhtin, A.; and Smith, N.~A. 2024.
\newblock RewardBench: Evaluating Reward Models for Language Systems.
\newblock \emph{arXiv preprint arXiv:2403.13787}.

\bibitem[{Lee et~al.(2023{\natexlab{a}})Lee, Phatale, Mansoor, and
  et~al.}]{lee2023rlaif}
Lee, H.; Phatale, S.; Mansoor, H.; and et~al. 2023{\natexlab{a}}.
\newblock RLAIF vs. RLHF: Scaling Reinforcement Learning from Human Feedback
  with AI Feedback.
\newblock \emph{arXiv preprint arXiv:2309.00267}.

\bibitem[{Lee et~al.(2023{\natexlab{b}})Lee, Phatale, Mansoor
  et~al.}]{rlaif_2023}
Lee, H.; Phatale, S.; Mansoor, H.; et~al. 2023{\natexlab{b}}.
\newblock RLAIF vs. RLHF: Scaling Reinforcement Learning from Human Feedback
  with AI Feedback.
\newblock In \emph{ICML 2024 Proceedings}.

\bibitem[{Li et~al.(2025)Li, He, Zu, Li, Shi, Xie, and Zhang}]{li2025multi}
Li, X.; He, Y.; Zu, S.; Li, Z.; Shi, T.; Xie, Y.; and Zhang, K. 2025.
\newblock Multi-Modal Large Language Model with RAG Strategies in Soccer
  Commentary Generation.
\newblock In \emph{2025 IEEE/CVF Winter Conference on Applications of Computer
  Vision (WACV)}, 6197--6206. IEEE.

\bibitem[{Liang and colleagues(2024)}]{preference2024multi}
Liang, Y.; and colleagues. 2024.
\newblock Preference Optimization with Multi-Sample Comparisons.
\newblock \emph{arXiv preprint arXiv:2410.12138}.

\bibitem[{Lin, Hilton, and Evans(2021)}]{lin2021truthfulqa}
Lin, S.; Hilton, J.; and Evans, O. 2021.
\newblock TruthfulQA: Measuring How Models Mimic Human Falsehoods.
\newblock \emph{arXiv preprint arXiv:2109.07958}.

\bibitem[{Lin et~al.(2024)Lin, Seto, ter Hoeve et~al.}]{dporm_limit_2024}
Lin, Y.; Seto, S.; ter Hoeve, M.; et~al. 2024.
\newblock On the Limited Generalization Capability of the Implicit Reward Model
  Induced by Direct Preference Optimization.
\newblock \emph{arXiv preprint arXiv:2409.03650}.

\bibitem[{Liu, Xu et~al.(2024)}]{xu2024aot}
Liu, B.-X.; Xu, J.; et~al. 2024.
\newblock Algorithm of Thoughts: Enhancing Exploration of Ideas in Large
  Language Models.
\newblock arXiv:2308.10379.

\bibitem[{Liu et~al.(2024{\natexlab{a}})Liu, Zeng, Xiao, He, and
  Yan}]{liu2024skywork}
Liu, C.; Zeng, L.; Xiao, Y.; He, J.; and Yan, R. 2024{\natexlab{a}}.
\newblock Skywork-Reward: Bag of Tricks for Reward Modeling in LLMs.
\newblock \emph{arXiv preprint arXiv:2410.18451}.

\bibitem[{Liu et~al.(2024{\natexlab{b}})Liu, Qin, Wu, Shen
  et~al.}]{liu2024lipo}
Liu, T.; Qin, Z.; Wu, J.; Shen, J.; et~al. 2024{\natexlab{b}}.
\newblock LiPO: Listwise Preference Optimization through Learning-to-Rank.
\newblock arXiv:2402.01878.

\bibitem[{Liu et~al.(2023)Liu, Zheng, Huang, Tang, Jiang, and Pan}]{liu2023}
Liu, Z.; Zheng, Y.; Huang, A.; Tang, T.; Jiang, M.; and Pan, L. 2023.
\newblock G-Eval: NLG Evaluation Using GPT-4 with Better Human Alignment.
\newblock \emph{arXiv preprint arXiv:2303.16634}.

\bibitem[{Maity and Saikia(2025)}]{maity2025}
Maity, S.; and Saikia, M.~J. 2025.
\newblock Large Language Models in Healthcare and Medical Applications: A
  Review.
\newblock \emph{Bioengineering}, 12(6): 631.

\bibitem[{Mnih et~al.(2016)Mnih, Badia, Mirza, and et~al.}]{mnih2016a3c}
Mnih, V.; Badia, A.~P.; Mirza, M.; and et~al. 2016.
\newblock Asynchronous Methods for Deep Reinforcement Learning.
\newblock In \emph{ICML}.
\newblock ArXiv:1602.01783.

\bibitem[{Muldrew et~al.(2024)Muldrew, Hayes, Zhang, and Barber}]{muldrew2024}
Muldrew, W.; Hayes, P.; Zhang, M.; and Barber, D. 2024.
\newblock Active Preference Learning for Large Language Models.
\newblock \emph{arXiv preprint arXiv:2402.08114}.

\bibitem[{Ouyang et~al.(2022)Ouyang, Wu, Jiang, Almeida, Wainwright
  et~al.}]{ouyang2022training}
Ouyang, L.; Wu, J.; Jiang, X.; Almeida, D.; Wainwright, C.~L.; et~al. 2022.
\newblock Training Language Models to Follow Instructions with Human Feedback.
\newblock arXiv:2203.02155.

\bibitem[{Paul(2024)}]{meta_reuters_2024}
Paul, K. 2024.
\newblock Meta releases AI model that can check other AI models' work.
\newblock \emph{Reuters}.
\newblock Oct 18 2024, Technology section.

\bibitem[{Plackett(1975)}]{plackett1975}
Plackett, R.~L. 1975.
\newblock The Analysis of Permutations.
\newblock \emph{Applied Statistics}, 24: 193--202.

\bibitem[{Rafailov et~al.(2023)Rafailov, Sharma, Mitchell et~al.}]{dpo_2023}
Rafailov, R.; Sharma, A.; Mitchell, E.; et~al. 2023.
\newblock Direct Preference Optimization: Your Language Model is Secretly a
  Reward Model.
\newblock \emph{Advances in Neural Information Processing Systems}.

\bibitem[{Ranaldi and Freitas(2024)}]{selfrefine2024}
Ranaldi, L.; and Freitas, A. 2024.
\newblock Self-Refine Instruction-Tuning for Aligning Reasoning in Language
  Models.
\newblock \emph{arXiv preprint arXiv:2405.00402}.

\bibitem[{Rao et~al.(2025)Rao, Zeng, Zhao, and Miao}]{rao2025survey}
Rao, H.; Zeng, M.; Zhao, X.; and Miao, C. 2025.
\newblock A survey of artificial intelligence in gait-based neurodegenerative
  disease diagnosis.
\newblock \emph{Neurocomputing}, 626: 129533.

\bibitem[{Rozi and contributors(2023)}]{vllm2023}
Rozi, E.; and contributors. 2023.
\newblock vLLM: Fast and Memory-Efficient LLM Serving.
\newblock \url{https://github.com/vllm-project/vllm}.

\bibitem[{Schulman et~al.(2017)Schulman, Wolski, Dhariwal, Radford, and
  Klimov}]{schulman2017ppo}
Schulman, J.; Wolski, F.; Dhariwal, P.; Radford, A.; and Klimov, O. 2017.
\newblock Proximal Policy Optimization Algorithms.
\newblock \emph{arXiv preprint arXiv:1707.06347}.

\bibitem[{Shao et~al.(2024)Shao, Wang, Zhu, Xu, Song, Bi, Zhang, Zhang, Li, Wu
  et~al.}]{shao2024deepseekmath}
Shao, Z.; Wang, P.; Zhu, Q.; Xu, R.; Song, J.; Bi, X.; Zhang, H.; Zhang, M.;
  Li, Y.; Wu, Y.; et~al. 2024.
\newblock Deepseekmath: Pushing the limits of mathematical reasoning in open
  language models.
\newblock \emph{arXiv preprint arXiv:2402.03300}.

\bibitem[{Song et~al.(2023)Song, Yu, Li, and et~al.}]{song2023pro}
Song, F.; Yu, B.; Li, M.; and et~al. 2023.
\newblock Preference Ranking Optimization for Human Alignment.
\newblock \emph{arXiv preprint arXiv:2306.17492}.

\bibitem[{Stein and colleagues(2025)}]{stein2025uncertainty}
Stein, A.; and colleagues. 2025.
\newblock Measuring Language Model Uncertainty with Internal Entropy.
\newblock In \emph{International Conference on Learning Representations}.

\bibitem[{Stiennon et~al.(2020)Stiennon, Ouyang, Wu, and et~al.}]{stiennon2020}
Stiennon, N.; Ouyang, L.; Wu, J.; and et~al. 2020.
\newblock Learning to Summarize from Human Feedback.
\newblock In \emph{NeurIPS}.
\newblock ArXiv:2009.01325.

\bibitem[{Stolfo et~al.(2024)Stolfo, Wu, Gurnee, and
  et~al.}]{stolfo2024entropy}
Stolfo, A.; Wu, B.; Gurnee, W.; and et~al. 2024.
\newblock Confidence Regulation Neurons in Language Models.
\newblock \emph{arXiv preprint arXiv:2406.16254}.

\bibitem[{Sun et~al.(2023)Sun, Yan, Yu, and Artzi}]{sun2023selfalign}
Sun, Z.; Yan, X.; Yu, T.; and Artzi, Y. 2023.
\newblock Principle-Driven Self-Alignment of Language Models from Scratch with
  Minimal Human Supervision.
\newblock arXiv:2305.03047.

\bibitem[{Tang et~al.(2021)Tang, Zhang, Yu, Turner, Derr, Wang, and
  Ntoutsi}]{tang2021interpretable}
Tang, X.; Zhang, W.; Yu, Y.; Turner, K.; Derr, T.; Wang, M.; and Ntoutsi, E.
  2021.
\newblock Interpretable visual understanding with cognitive attention network.
\newblock In \emph{International Conference on Artificial Neural Networks},
  555--568. Springer.

\bibitem[{Vanlioglu(2025)}]{vanlioglu2025}
Vanlioglu, A. 2025.
\newblock Entropy-Guided Sequence Weighting for Efficient Exploration in
  RL-Based LLM Fine-Tuning.
\newblock \emph{arXiv preprint arXiv:2503.22456}.

\bibitem[{Wan et~al.(2023)Wan, Sun, Dai, Ar{\i}k, and Pfister}]{wan2023}
Wan, X.; Sun, R.; Dai, H.; Ar{\i}k, S.~O.; and Pfister, T. 2023.
\newblock Better Zero-Shot Reasoning with Self-Adaptive Prompting.
\newblock In \emph{Findings of ACL}.

\bibitem[{Wang et~al.(2025{\natexlab{a}})Wang, Gan, Huo et~al.}]{gram_2025}
Wang, C.; Gan, Y.; Huo, Y.; et~al. 2025{\natexlab{a}}.
\newblock GRAM: A Generative Foundation Reward Model for Reward Generalization.
\newblock In \emph{ICML 2025}.

\bibitem[{Wang, Huang et~al.(2024)}]{huang2024online}
Wang, H.; Huang, F.; et~al. 2024.
\newblock Self-Improving Efficient Online Alignment of Large Language Models.
\newblock arXiv:2406.15567.

\bibitem[{Wang et~al.(2025{\natexlab{b}})Wang, He, Zhong, Song, Su, Feng, He,
  Zhu, Yuan, Lu et~al.}]{wang2025twin}
Wang, J.; He, Y.; Zhong, Y.; Song, X.; Su, J.; Feng, Y.; He, H.; Zhu, W.; Yuan,
  X.; Lu, K.; et~al. 2025{\natexlab{b}}.
\newblock Twin Co-Adaptive Dialogue for Progressive Image Generation.
\newblock \emph{arXiv preprint arXiv:2504.14868}.

\bibitem[{Wang et~al.(2024)Wang, Zhang, He, Zhang, Song, Shi, Li, Xu, Wu, Yi
  et~al.}]{wang2024enhancing}
Wang, J.; Zhang, Z.; He, Y.; Zhang, Z.; Song, Y.; Shi, T.; Li, Y.; Xu, H.; Wu,
  K.; Yi, X.; et~al. 2024.
\newblock Enhancing Code LLMs with Reinforcement Learning in Code Generation: A
  Survey.
\newblock \emph{arXiv preprint arXiv:2412.20367}.

\bibitem[{Wang et~al.(2025{\natexlab{c}})Wang, Yu, Gao, Zheng, Liu
  et~al.}]{wang2025fork}
Wang, S.; Yu, L.; Gao, C.; Zheng, C.; Liu, S.; et~al. 2025{\natexlab{c}}.
\newblock Beyond the 80/20 Rule: High-Entropy Minority Tokens Drive Effective
  Reinforcement Learning for LLM Reasoning.
\newblock In \emph{ACL}.
\newblock ArXiv:2506.01939.

\bibitem[{Wang et~al.(2022)Wang, Wei, Schuurmans et~al.}]{wang2022sc}
Wang, X.; Wei, J.; Schuurmans, D.; et~al. 2022.
\newblock Self-Consistency Improves Chain of Thought Reasoning in Language
  Models.
\newblock arXiv:2203.11171.

\bibitem[{Wei et~al.(2022)Wei, Wang, Schuurmans et~al.}]{wei2022cot}
Wei, J.; Wang, X.; Schuurmans, D.; et~al. 2022.
\newblock Chain-of-Thought Prompting Elicits Reasoning in Large Language
  Models.
\newblock arXiv:2201.11903.

\bibitem[{Weng(2018)}]{weng2018policygrad}
Weng, L. 2018.
\newblock Policy Gradient Algorithms.
\newblock \url{https://lilianweng.github.io/posts/2018-04-08-policy-gradient/}.

\bibitem[{Williams(1992)}]{williams1992}
Williams, R.~J. 1992.
\newblock Simple Statistical Gradient-Following Algorithms for Connectionist
  Reinforcement Learning.
\newblock \emph{Machine Learning}, 8(3--4): 229--256.

\bibitem[{Wu et~al.(2021)Wu, Ouyang, Ziegler, and et~al.}]{wu2021}
Wu, J.; Ouyang, L.; Ziegler, D.~M.; and et~al. 2021.
\newblock Recursively Summarizing Books with Human Feedback.
\newblock \emph{arXiv preprint arXiv:2109.10862}.

\bibitem[{Wu et~al.(2024{\natexlab{a}})Wu, Yuan, Golovneva
  et~al.}]{meta_rewarding_2024}
Wu, T.; Yuan, W.; Golovneva, O.; et~al. 2024{\natexlab{a}}.
\newblock Meta-Rewarding Language Models: Self-Improving Alignment with
  LLM-as-a-Meta-Judge.
\newblock \emph{arXiv preprint arXiv:2407.19594}.

\bibitem[{Wu et~al.(2024{\natexlab{b}})Wu, Xu, Cui, Zhan, Zhu, and
  Feng}]{wu2024}
Wu, Z.; Xu, B.; Cui, R.; Zhan, M.; Zhu, X.; and Feng, L. 2024{\natexlab{b}}.
\newblock Rethinking Chain-of-Thought from the Perspective of Self-Training.
\newblock \emph{arXiv preprint arXiv:2412.10827}.

\bibitem[{Xin et~al.(2024)Xin, Luo, Zhou, Du, Liu, Fan, Li, and
  Du}]{xin2024parameter}
Xin, Y.; Luo, S.; Zhou, H.; Du, J.; Liu, X.; Fan, Y.; Li, Q.; and Du, Y. 2024.
\newblock Parameter-efficient fine-tuning for pre-trained vision models: A
  survey.
\newblock \emph{arXiv preprint arXiv:2402.02242}.

\bibitem[{Yang, Zhao, and Shen(2025)}]{yang2025ehrstruct}
Yang, X.; Zhao, X.; and Shen, Z. 2025.
\newblock EHRStruct: A Comprehensive Benchmark Framework for Evaluating Large
  Language Models on Structured Electronic Health Record Tasks.
\newblock \emph{arXiv preprint arXiv:2511.08206}.

\bibitem[{Yu et~al.(2025)Yu, Lin, Wu, and et~al.}]{yu2025}
Yu, T.; Lin, T.-E.; Wu, Y.; and et~al. 2025.
\newblock Diverse AI Feedback for Large Language Model Alignment.
\newblock \emph{Transactions of the Association for Computational Linguistics},
  13: 392--407.

\bibitem[{Zhang et~al.(2024)Zhang, Wang, Jin et~al.}]{proto_rm_2024}
Zhang, J.; Wang, X.; Jin, Y.; et~al. 2024.
\newblock Prototypical Reward Network for Data-Efficient RLHF.
\newblock In \emph{ACL 2024}.
\newblock ArXiv:2406.06606.

\bibitem[{Zhang, Liu et~al.(2024)}]{dposurvey2024}
Zhang, T.; Liu, J.; et~al. 2024.
\newblock A Comprehensive Survey of Direct Preference Optimization.
\newblock \emph{arXiv preprint arXiv:2410.15595}.

\bibitem[{Zhao, Dang, and Grover(2024)}]{zhao2024gpo}
Zhao, S.; Dang, J.; and Grover, A. 2024.
\newblock Group Preference Optimization: Few-Shot Alignment of Large Language
  Models.
\newblock In \emph{ICLR}.
\newblock ArXiv:2310.11523.

\bibitem[{Zhao et~al.(2021)Zhao, Chen, Xing, and Miao}]{zhao2021brain}
Zhao, X.; Chen, H.; Xing, Z.; and Miao, C. 2021.
\newblock Brain-inspired search engine assistant based on knowledge graph.
\newblock \emph{IEEE Transactions on Neural Networks and Learning Systems},
  34(8): 4386--4400.

\bibitem[{Zhao et~al.(2025{\natexlab{a}})Zhao, Liu, Yang, and
  Miao}]{zhao2025medrag}
Zhao, X.; Liu, S.; Yang, S.-Y.; and Miao, C. 2025{\natexlab{a}}.
\newblock Medrag: Enhancing retrieval-augmented generation with knowledge
  graph-elicited reasoning for healthcare copilot.
\newblock In \emph{Proceedings of the ACM on Web Conference 2025}, 4442--4457.

\bibitem[{Zhao et~al.(2025{\natexlab{b}})Zhao, Liu, Yang, and
  Miao}]{zhao2025smart}
Zhao, X.; Liu, S.; Yang, S.-Y.; and Miao, C. 2025{\natexlab{b}}.
\newblock A smart multimodal healthcare copilot with powerful llm reasoning.
\newblock \emph{arXiv preprint arXiv:2506.02470}.

\bibitem[{Zhao, Bai, and Zhao(2025)}]{zhao2025gfriend}
Zhao, Y.; Bai, H.; and Zhao, X. 2025.
\newblock GFRIEND: Generative Few-shot Reward Inference through EfficieNt DPO.
\newblock \emph{arXiv preprint arXiv:2506.08965}.

\bibitem[{Zheng et~al.(2023)Zheng, Zou, Chen et~al.}]{zheng2023rlaif}
Zheng, H.; Zou, A.; Chen, E.; et~al. 2023.
\newblock RLAIF vs.\ RLHF: Scaling Reinforcement Learning from Human Feedback
  without Humans.
\newblock arXiv:2309.00267.

\bibitem[{Zhou et~al.(2024{\natexlab{a}})Zhou, Zheng, Wang
  et~al.}]{rm_benchmark_2024}
Zhou, E.; Zheng, G.; Wang, B.; et~al. 2024{\natexlab{a}}.
\newblock RMB: Comprehensively Benchmarking Reward Models in LLM Alignment.
\newblock \emph{arXiv preprint arXiv:2410.09893}.

\bibitem[{Zhou et~al.(2025{\natexlab{a}})}]{zhou2025self}
Zhou, X.; et~al. 2025{\natexlab{a}}.
\newblock Self-Consistency of Internal Reward Models Improves Alignment.
\newblock \emph{arXiv preprint arXiv:2502.08922}.

\bibitem[{Zhou et~al.(2025{\natexlab{b}})Zhou, He, Su, Han, Jang, Bertasius,
  Bansal, and Yao}]{zhou2025reagent}
Zhou, Y.; He, Y.; Su, Y.; Han, S.; Jang, J.; Bertasius, G.; Bansal, M.; and
  Yao, H. 2025{\natexlab{b}}.
\newblock ReAgent-V: A Reward-Driven Multi-Agent Framework for Video
  Understanding.
\newblock \emph{arXiv preprint arXiv:2506.01300}.

\bibitem[{Zhou et~al.(2024{\natexlab{b}})Zhou, Zhang, Zhang, He, Wang, Shi, and
  Khamis}]{zhou2024human}
Zhou, Z.; Zhang, J.; Zhang, J.; He, Y.; Wang, B.; Shi, T.; and Khamis, A.
  2024{\natexlab{b}}.
\newblock Human-centric reward optimization for reinforcement learning-based
  automated driving using large language models.
\newblock \emph{arXiv preprint arXiv:2405.04135}.

\bibitem[{Ziegler et~al.(2019)Ziegler, Stiennon, Wu, and et~al.}]{ziegler2019}
Ziegler, D.~M.; Stiennon, N.; Wu, J.; and et~al. 2019.
\newblock Fine-Tuning Language Models from Human Preferences.
\newblock \emph{arXiv preprint arXiv:1909.08593}.

\end{thebibliography}

\end{document}